\documentclass[lettersize,journal]{IEEEtran}
\usepackage{amsmath,amsfonts}
\usepackage{algorithmic}
\usepackage{algorithm}
\usepackage{array}
\usepackage{subcaption}  % Add this line in the preamble
\usepackage{textcomp}
\usepackage{stfloats}
\usepackage{url}
\usepackage{verbatim}
\usepackage{graphicx}
\usepackage{cite}
\usepackage{hyperref}
\usepackage{orcidlink}
\usepackage{dirtytalk}
\usepackage{xcolor}
\usepackage{soul}
\usepackage{csquotes}

\begin{document}

\title{Enabling Advanced Land Cover Analytics: An Integrated Data Extraction Pipeline for Predictive Modeling with the Dynamic World Dataset}

\author{Victor Radermecker, Andrea Zanon, Nancy Thomas, Annita Vapsi, \\ Saba Rahimi, Rama Ramakrishnan, Daniel Borrajo
        % <-this % stops a space
\thanks{Manuscript received August 15 , 2024; revised July 14, 2025. \textit{(Victor Radermecker and Andrea Zanon are co-first authors.) (Corresponding authors: Victor Radermecker, Andrea Zanon.)}} 
\thanks{Victor Radermecker, Andrea Zanon, and Rama Ramakrishnan are with MIT Sloan School of Management and Operations Research Center, Massachusetts Institute of Technology, 
Cambridge, MA 02139 USA (e-mail: victor.radermecker@gmail.com, azanon@alum.mit.edu, ramar@mit.edu).}
\thanks{Nancy Thomas, Annita Vapsi, Saba Rahimi, and Daniel Borrajo are with JP Morgan, New York, NY 10017 USA (e-mail: nancy.thomas@jpmchase.com, annita.vapsi@jpmchase.com, saba.rahimi@jpmorgan.com, daniel.borrajo@jpmchase.com).}
}

% \markboth{Journal of Selected Topics in Applied Earth Observations and Remote Sensing,~Vol.~xx, No.~xx, xxxx~xxxx}%


\maketitle

\begin{abstract}
Understanding land cover holds considerable potential for a myriad of practical applications, particularly as data accessibility transitions from being exclusive to governmental and commercial entities to now including the broader research community. Nevertheless, although the data is accessible to any community member interested in exploration, there exists a formidable learning curve and no standardized process for accessing, pre-processing, and leveraging the data for subsequent tasks. In this study, we democratize this data by presenting a flexible and efficient end to end pipeline for working with the Dynamic World dataset, a cutting-edge near-real-time land use/land cover (LULC) dataset. This includes a pre-processing and representation framework which tackles noise removal, efficient extraction of large amounts of data, and re-representation of LULC data in a format well suited for several downstream tasks. To demonstrate the power of our pipeline, we use it to extract data for an urbanization prediction problem and build a suite of machine learning models with excellent performance. This task is easily generalizable to the prediction of any type of land cover and our pipeline is also compatible with a series of other downstream tasks. \\ To facilitate further research and validation, all code and data used in this study are made available as open source in the following \href{https://github.com/victor-radermecker/AdvancedLandCoverAnalytics-Pipeline}{GitHub repository}.
\end{abstract}


\vfill\null
% \columnbreak

\section{Introduction}
\IEEEpubidadjcol
\IEEEPARstart{R}{emote} sensing has enabled large scale land cover analysis, allowing researchers to explore phenomena such as deforestation, agricultural expansion, and the rapid pace of urbanization. For example, it has been used to detect illegal farms in the Amazon rain forest \cite{Stanford2024}, understand urban expansion patterns for public transportation development \cite{Karimi2024Mar}, and assess land use's impact on natural landscapes \cite{Shuangao2021Feb}. Insights from these analyses are vital for informed decision-making in environmental conservation, urban planning, and sustainable development. Historically, the construction of LULC datasets relied heavily on human effort, often resulting in time-consuming processes with limited updates. Today, the explosion of available satellite imagery, combined with advancements in computational resources and artificial intelligence, particularly in computer vision, has revolutionized this domain. In this work, we use the Dynamic World dataset which, as a near real-time 10m resolution LULC dataset, offers more accurate and granular segmentation in both temporal and spatial dimensions~\cite{Brown2022Jun}. 

Despite the increase in availability of remote sensing data and the steep learning curve associated with its use, there is not an end-to-end pipeline for extracting, pre-processing, and representing the data so that it can be easily used in a variety downstream tasks. Rather, many works perform this process in a one-off basis, which is not efficient, necessitates familiarity with remote sensing data, limits comparability between methods, and results in task specific datasets. ~\cite{rolf2024mission}
posits that satellite data should be treated as a distinct modality and therefore should be handled differently than standard image data with respect to pre-processing and modeling. We agree with this assertion and offer our solution as a system for handling this modality.

As demonstration of our LULC data pipeline, this paper presents a future LULC prediction model fit on data extracted and represented using the pipeline. In particular, we focus on urbanization prediction, although the task could be easily applied to any land cover type. We select this downstream task due to its relevance and nontriviality as a critical advancement in LULC analysis lies in predicting future land cover changes. This paper presents a model which successfully captures the dynamics of changing land cover and forecasts urbanization even in regions which are urbanizing quickly using a combined XGBoost-ConvLSTM model, called XGCLM. The results underscore the success of our data processing pipeline in extracting signal from the LULC data and representing the data so that it can be easily used alongside downstream machine learning models.

\section{Literature Review}
\subsection{Land Use Land Cover Datasets}
There are several open source LULC datasets, each with their own benefits and drawbacks. In this section, we will introduce a few of the most notable high resolution datasets.

The first high resolution global LULC dataset is GlobeLand30, which is available at irregular frequencies from 2000 to 2010~\cite{globeland30}. It has 30 meter resolution and classifies land into ten classes using a novel pixel-object-knowledge-based (POK-based) classification approach on top of Landsat imagery. The restricted timeline of GlobeLand30's availability and lack of recent updates make it suboptimal for our use case.

In 2021, the Dynamic World dataset was introduced as a collaboration between Google and the World Resources Institute~\cite{Brown2022Jun}. It supplies trustworthy LULC labels and therefore serves as a valuable asset for estimating urbanization rates without the need to directly process \say{raw} satellite images. These labels also help to establish consistency of ground truth LULC labels, allowing researchers to focus on downstream tasks and easily compare their results with other works. 

In contrast to conventional methods, the Dynamic World dataset employs cutting-edge segmentation techniques on Sentinel-2 Top of Atmosphere imagery to define 10 bands, 9 of which hold the estimated probability that a pixel is completely covered by a particular class, and a final band which holds the index of the band with the highest estimated probability. These segmentation techniques allow for the continuous generation of land cover updates with little overhead. The 9 available labels include water, trees, grass, flooded vegetation, crops, shrub and scrub, built, bare, snow and ice. In our downstream case study, we focus on the built band, but our framework is generalizable and could be applied to any band without changing the methodology. Through integration with the European Commission's Copernicus Programme and Google Earth Engine, Dynamic World achieves an impressive 10m resolution, ensuring the delivery of highly precise outcomes~\cite{gee, Brown2022Jun}.

One possible alternative for Dynamic World is the National Land Cover Database (NLCD), released by the U.S. Geological Survey (USGS) in association with the Multi-Resolution Land Characteristics (MRLC) Consortium.
The NLCD systematically classifies and characterizes land cover throughout the United States into 16 land cover categories like forests, urban areas, agricultural fields, water bodies, and more.
The outcome is a multi-class dataset where each pixel is ascribed a specific land cover class \cite{BibEntry2024Apr}.

Many works in urbanization forecasting do not leverage the existing LULC labels available in open source datasets, such as the Dynamic World dataset, and instead train their own models to first predict the LULC labels ~\cite{nkt1, nkt2, nkt4, nkt5, nkt7, nkt8, nkt10, nkt12, nkt13, nkt14, nkt15, nkt16, nkt17, nkt18}. The use of our pipeline in conjunction with open source LULC datasets should eliminate the need for custom LULC labelling. 
It should be noted that LULC labels extracted from such datasets, while of high quality, are not ground truth since they are generated by deep learning classification algorithms. As a consequence, we do not dismiss alternative approaches that may provide higher accuracy or robustness. Rather, our use of open source datasets reflects a practical concern: many teams lack the resources (computational or time) to generate large-scale, representative LULC labels, making datasets like Dynamic World a valuable baseline for global-scale applications. We also note that there may be other open source LULC datasets which outperform Dynamic World in some regions and time periods. Our pipeline could be easily amended to include these datasets, but we leave this line of work for future studies.

\subsection{Land Use Land Cover Dataset Processing Pipeline}

Despite the difficulty and steep learning curve associated with working with satellite imagery, to the extent of our knowledge, there does not exist another open source pipeline like ours which includes extraction, pre-processing, and representation of LULC data. Many works prefer to use their own LULC labeling rather than using datasets like Dynamic World, and therefore rely on custom downstream pre-processing implementations. 

For instance, some methods ~\cite{nkt2} rely on georeferencing imagery and image-to-image registration to align data into a common coordinate system, followed by the use of commercial software (e.g., eCognition Developer) for object-based image analysis to derive LULC classifications. In contrast, our pipeline eliminates the need for these additional steps by directly extracting images in a consistent coordinate system. Additionally, our approach is entirely open-source, built in Python, and does not depend on proprietary systems.

Other methods employ supervised classification approaches, such as using historical imagery and manually digitized polygons as training data for classifiers ~\cite{nkt4, nkt16}. These approaches rely on the assumption that classifiers trained on smaller, localized datasets will generalize effectively to broader, unseen regions—a limitation our pipeline avoids. On the other hand, unsupervised methods, like the segmentation-based approach in ~\cite{nkt18}, utilize Convolutional Neural Networks (CNNs) for feature clustering but often require manual corrections to remove residual noise. This manual intervention significantly limits scalability, especially when applied to large geographic regions. 

There is a package, \href{https://geemap.org/}{GeeMap}, which offers tools for satellite data analysis and visualization. As we will discuss, we have build upon and extended this package as its initial implementation does not include the end to end pipeline offered in our work.

\subsection{Urbanization Monitoring and Prediction}\label{sec:urbanlit}

As previously mentioned, we have chosen as our downstream task the analysis and prediction of urbanization as a subset of the broader study of land cover monitoring. Therefore, we focus on built land, but the same analysis could be applied to any other type of land labeled in a LULC dataset (eg. crops, water, etc.). In our review, we hope to show both the relevance and nontriviality of this task. There have been several works in this space, however, as we will discuss in Section ~\ref{sec:methods}, our streamlined pipeline for processing LULC data allows us to achieve competitive results on the task of urbanization prediction.

Recent advancements in urban sprawl analysis have focused on understanding the multifaceted impact of urban growth on the environment and societal structures. These studies \cite{de2023methodological,  csahin2022investigation} contribute to a nuanced understanding of urbanization that includes environmental and infrastructural considerations. The integration of big data and machine learning for urban change detection analysis signifies a pivotal shift towards data-driven urban planning and management \cite{praveen2023urban}.

Many works in urbanization monitoring and prediction have chosen to focus their study on a particular geographic region ~\cite{nkt2, nkt4, nkt7, nkt13, nkt14, nkt15, nkt16, nkt19}, in some cases augmenting satellite image data with region specific information ~\cite{nkt5, nkt6, nkt11, nkt15, nkt16}. While we will demonstrate our results on a case study region, it was chosen arbitrarily and our method is geographically generalizable as it doesn't necessitate the inclusion of external sources.  Some works have focused on urbanization at a macro scale, tracking the urbanization of an entire city as a whole ~\cite{nkt1, nkt3, nkt15, nkt17}. These works employ strategies such as entropy to study the dynamics of city expansion ~\cite{nkt1, nkt15}. We instead conduct our case study at a more granular level, tracking urbanization at the level of small geographic areas. We focus more on representing current urbanization and predicting future urbanization, rather than interpretable modeling of an area’s urbanization profile as a whole, as our goal is to demonstrate the efficacy of our data processing pipeline rather than to obtain specific results on a particular area.

In terms of the methods behind urbanization forecasting, there seem to be two primary approaches in the literature: Markov Chains with Cellular Automata (MC-CA) \cite{nkt2, nkt4, nkt11, nkt14, nkt16, nkt19} and deep learning~\cite{nkt7, nkt18}. The general framework of MC-CA is as follows. First, a model is trained to detect LULC labels from Landsat images. Next, a Markov Chain model is used to estimate transition probabilities of land cover. Finally, a Cellular Automata model is used alongside the estimated transition probabilities and, if available, some region specific data such as road maps or elevation maps, to model changes in land cover in a region and forecast future urbanization. 

On the deep learning side,  convolutional neural networks, deeply supervised attention metric-based networks, and dual-task Siamese networks have been applied for urban change detection ~\cite{daudt2018urban, shi2021deeply, hafner2022urban}.  Some works use video prediction as a means for forecasting urbanization \cite{nkt7, nkt18}. These works view a time series of satellite images as a video and use video prediction modeling to predict future states of urbanization. \cite{nkt7}  predicts future urbanization in five cities in the Middle East and North Africa, focusing on large scale urbanization pertaining to an entire city, rather than fine grained urbanization on small geographic areas.~\cite{nkt18} uses an unsupervised image segmentation method to identify built regions and  a ConvLSTM to predict future urbanization.

\subsection{Video Prediction}

In our exemplary case study, we are focusing not only on analyzing but also predicting land cover and land usage. In order to leverage both the temporal and spatial nature of our data, we explored video prediction models. In particular, we used the Convolutional LSTM (ConvLSTM) network, which was originally used for precipitation nowcasting~\cite{convlstm}. The ConvLSTM network combines the strengths of convolutional neural networks for extracting spatial features and long short-term memory (LSTM) networks for capturing temporal dependencies to effectively model spatio-temporal patterns. Specifically, the ConvLSTM replaces the fully connected layers in a standard LSTM with convolutional layers, allowing the model to learn spatially-varying temporal relationships directly from the input data. This enables the ConvLSTM to extract relevant spatial features and understand the evolution of these features over time, making it well-suited for future LULC prediction which requires modeling complex spatio-temporal patterns.

\section{Materials and Methods}
\label{sec:methods}

\subsection{Dataset\label{Dataset}}

For this work, we have chosen to use the Dynamic World dataset. Unlike GlobeLand30, Dynamic World offers near real time data and is currently available up to the current date. While NLCD offers manual labeling and a historical span that extends over several years, its limited update frequency of 2 to 3 years (as observed in the last four updates in 2013, 2016, 2019, 2021) poses challenges for continuous monitoring. In contrast, Dynamic World, with near real-time updates and global coverage, provides a more dynamic and responsive representation of changing land cover conditions. This makes it preferable for tasks requiring frequent monitoring, like ours.

\subsection{Methodology \label{methodology}}

In this paper, we introduce a pipeline designed to efficiently extract, pre-process, and re-represent LULC remote sensing data. In addition to analysis of current land cover and historical changes, this pipeline can be directly used alongside machine learning models to perform a series of downstream tasks. We will demonstrate one such downstream task, which is the prediction of future urbanization.

\subsubsection{Data Extraction and pre-processing}

While Dyamic World offers high quality LULC labels, there still does not exist an established standard for pre-processing and representing LULC data. The major focus of our work is to develop a data extraction and image pre-processing pipeline to address the magnitude inherent to satellite data and the potential noise in each image. In this section, we offer an explanation of our pipeline — a novel addition to the literature and one that we were missing when beginning the work.

Altough our pipeline was developed using images extracted from the Dynamic World dataset, it is not heavily dependent on it and could easily be amended to use a different source or include additional datasets.

\begin{figure*}[!t]
\centering
\resizebox{0.8\textwidth}{!}{%
\includegraphics{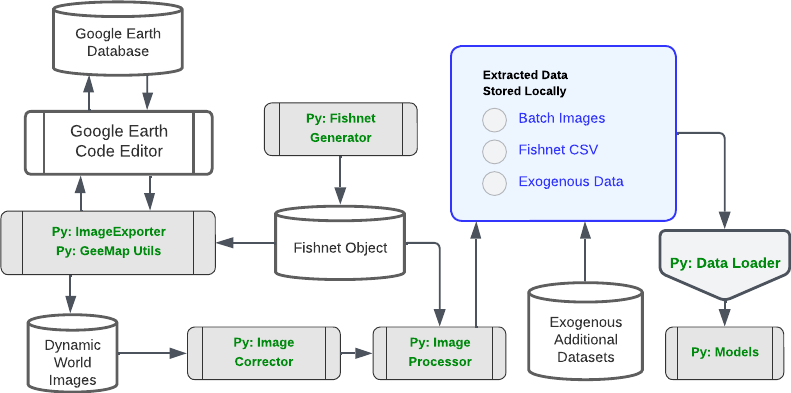}
}
\caption{The data processing flow begins with the Google Earth Database, which is queried using the Google Earth Code Editor. Python modules facilitate the extraction of images from the database and store them in a Dynamic World Images repository. A Fishnet Generator creates a grid for image analysis. After correction and processing, the resulting data, including images, a fishnet CSV, and additional data, are stored locally. This local repository then feeds into a Data Loader, which prepares the data for analysis using Python models. Additional data sources can be joined at fishnet granularity.}
\label{fig:extraction_pre-processing}
\end{figure*}

A diagram providing an overview of all implemented modules is provided in Figure \ref{fig:extraction_pre-processing}.

The entire region of interest is partitioned into a “fishnet”, a block of fixed-sized tiles, and for each tile the corresponding satellite image is extracted. The system's foundation revolves around this fishnet. 
The approach of partitioning a larger geographical region into smaller, uniformly sized areas offers distinct advantages in granularity, facilitating analysis at the desired level of detail as one can combine tiles for a broader perspective and analyze individual tiles for finer resolution. By selecting a subset of tiles, focus can be directed toward a specific region of interest, ensuring both concentration and consistency throughout the code as each spatial point is mapped to its corresponding tile. Additionally, this method helps overcome challenges related to data management, particularly with large-scale satellite imagery, by facilitating quicker exportation and loading of data. It also supports comparative analysis by allowing the extraction of features for regions of equal size, enabling comparisons between different areas based on these features. Furthermore, the features extracted from each region, including the images corresponding to the tiles, can be seamlessly integrated into machine learning models.

The fishnet approach does not compete with the existing Google Earth Engine (GEE) capabilities, but rather complements them. While GEE provides robust capabilities for selecting and processing specific regions, the fishnet approach addresses key challenges related to scalability, standardization, and ease of integration into machine learning workflows.
\begin{itemize}
    \item Scalability: GEE’s region-based processing works well for small areas, but faces challenges with large regions spanning multiple administrative units such as counties or zip codes due to their irregular size. The fishnet divides the area into uniform grid cells, enabling consistent granularity and seamless processing. Even using the Rectangle geometry within GEE does not work well; large areas cannot be easily exported, often necessitating manual segmentation. Although GEE’s built-in methods may be effective for extracting data from a few counties, significant modifications would need to be made to allow for extraction of data at scale. This is where the fishnet excels.
    \item Standardization: Machine learning models require input data to be consistent in size and format; the fishnet ensures that all extracted image patches are of equal dimensions, eliminating the need for additional resizing or cropping steps that would be necessary using administrative boundaries like counties or zip codes. Even using GEE's Rectangle geometry can result in non-uniform sizes, as the image dimensions depend on the absolute latitude of a region. For example, a square region of 0.1 degrees by 0.1 degrees in latitude and longitude would have different dimensions when exported as an image if it is near one of the poles versus near the equator. This irregularity is handled by the fishnet. There is a method within GEE that allows for tiled exporting of large areas, but generates web-friendly map tiles optimized for visualization, rather than data to be ingested by a machine learning model, unlike the fishnet we have proposed.
    \item Ease of Processing: The fishnet grid allows for systematic and parallelized extraction of data across the entire study region. This design facilitates streamlined integration with machine learning pipelines, where uniform inputs are critical for efficient training and inference. To the extent of our knowledge, there does not exist a built-in GEE function which offers extraction of very large scale regions into uniformly sized areas with a mapping between the extracted data and the tile location.
\end{itemize}

The fishnet outline is initialized through either the minimum and maximum latitude and longitude or by providing a shapefile, as well as a parameter specifying the size of each tile. The above pseudocode (Algorithm \ref{alg:generateFishnet}) outlines the process of generating a fishnet grid for spatial data analysis. 

\begin{algorithm*}[!t] %[H]
\caption{Generate Fishnet Grid.}\label{alg:generateFishnet}
\begin{algorithmic}
\STATE 
\STATE {\textsc{GenerateFishnet}}$(\text{boundingBox}, \text{tileSize})$
\STATE \hspace{0.5cm} \textbf{Input:} boundingBox (minLat, minLon, maxLat, maxLon), tileSize (meters)
\STATE \hspace{0.5cm} \textbf{Output:} Fishnet grid
\STATE 
\STATE \hspace{0.5cm} \textbf{Convert} boundingBox coordinates to meters: \texttt{topLeft, bottomRight} $\gets$ \texttt{ConvertToMeters(boundingBox)}
\STATE 
\STATE \hspace{0.5cm} \textbf{Calculate} number of tiles:
\STATE \hspace{1.0cm} $\texttt{numTilesX} \gets \texttt{ceil}((\texttt{bottomRight.x} - \texttt{topLeft.x}) / \texttt{tileSize})$
\STATE \hspace{1.0cm} $\texttt{numTilesY} \gets \texttt{ceil}((\texttt{topLeft.y} - \texttt{bottomRight.y}) / \texttt{tileSize})$
\STATE 
\STATE \hspace{0.5cm} \textbf{Initialize} empty list: \texttt{fishnet} $\gets$ []
\STATE 
\STATE \hspace{0.5cm} \textbf{For} $i = 0$ \textbf{to} $\texttt{numTilesX} - 1$ \textbf{do}
\STATE \hspace{1.0cm} \textbf{For} $j = 0$ \textbf{to} $\texttt{numTilesY} - 1$ \textbf{do}
\STATE \hspace{1.5cm} \textbf{Calculate} left coordinate of current tile: \texttt{tileLeft} $\gets$ \texttt{topLeft.x} $+ i \times \texttt{tileSize}$
\STATE \hspace{1.5cm} \textbf{Calculate} bottom coordinate of current tile: \texttt{tileBottom} $\gets$ \texttt{bottomRight.y} $+ j \times \texttt{tileSize}$
\STATE \hspace{1.5cm} \textbf{Convert} tile coordinates to latitude and longitude: \texttt{tileBoundingBox} $\gets$ \texttt{ConvertToLatLon( \\ \hspace{3cm} tileLeft, tileBottom, tileSize)}
\STATE \hspace{1.5cm} \textbf{Add} current tile's bounding box to \texttt{fishnet} list
\STATE 
\STATE \hspace{0.5cm} \textbf{Return} \texttt{fishnet}
\end{algorithmic}
\end{algorithm*}

As a practical way for converting degrees to meters and vice versa, which is necessary to correctly extract images from Google Earth Engine, we employ the Haversine formula. It calculates the great-circle distance between two points on the Earth's surface, considering their latitude and longitude coordinates. Despite the Earth’s non-perfect spherical shape, the Haversine formula, given by Equation \ref{eq:haversine}, is generally considered a good approximation for such use cases.

\begin{equation}
\label{eq:haversine}
\small
d = 2r \arcsin\left(\sqrt{\sin^2\left(\frac{\Delta \varphi}{2}\right) + \cos(\varphi_1) \cos(\varphi_2) \sin^2\left(\frac{\Delta \lambda}{2}\right)}\right)
\end{equation}

where:
\begin{itemize}
\item \( d \) is the distance between the two points,
\item \( r \) is the radius of the Earth,
\item \( \varphi_1, \varphi_2 \) are the latitudes of the two points in radians,
\item \( \Delta \varphi \) is the difference in latitudes,
\item \( \Delta \lambda \) is the difference in longitudes.
\end{itemize}

Due to the substantial volume of images, the development of an efficient method for extraction is necessary. To give an example, the state of Texas spans approximately 1300 km from north to south and 1250 km from east to west. Utilizing a high-resolution grid with each fishnet tile measuring 400 meters per side, the total is approximately 10 million tiles. It is clear that extracting each image individually would be impractical. However, there is a trade-off with processing time, as many smaller images are faster to process than few bigger ones. 
To process and analyze such large scale of images, we employ a batching approach. Considering a batch size of 64 km per side, around 400 batch images would be required to comprehensively cover Texas. Extracting 400 high dimension images is much easier than extracting millions of smaller ones; at the same time, pre-processing 400 images is faster than pre-processing a dozen of images representing the whole region. 

\begin{figure}[H]
\includegraphics[width=8.9 cm]{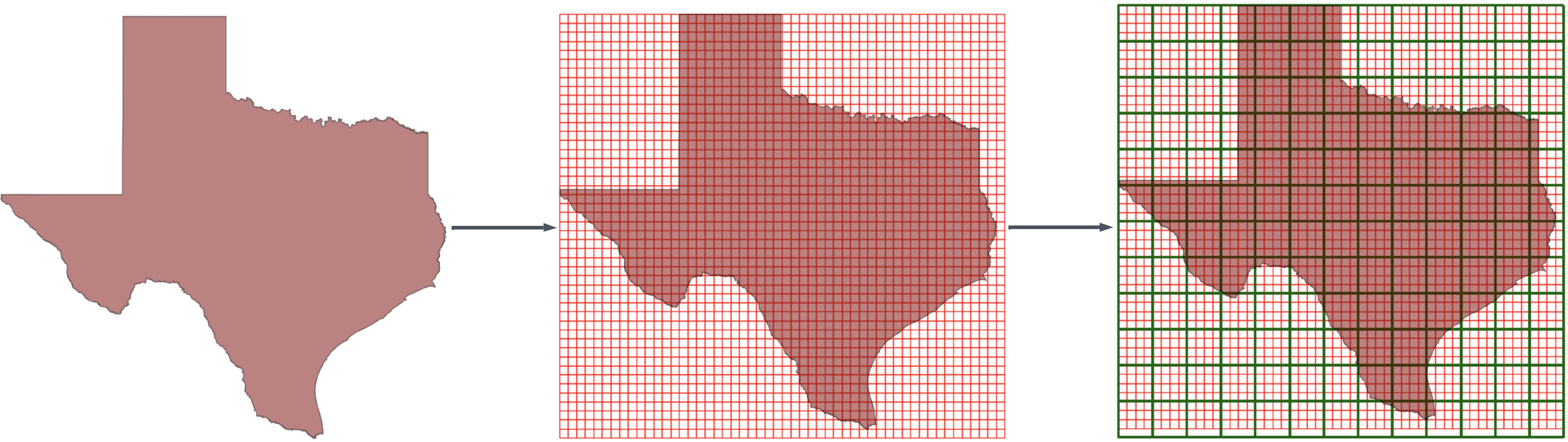}
\caption{The process begins by either obtaining a shapefile representation of the desired region, the state of Texas in this example, or by specifying the top-left and bottom-right latitude and longitude coordinates. Subsequently, we generate a fishnet grid, as explained in previous sections. Due to the large number of fishnet tiles, we opt to aggregate the tiles into batches before extracting them from Google Earth Engine, resulting in fewer but larger exported images. The diagram provided above depicts the batching process from left to right. The fishnet tiles are represented in red whereas the batches are represented in green.}
\label{fig:batching}
\end{figure}

Figure \ref{fig:batching} illustrates the batching process. This methodology allows for efficient data extraction for each fishnet tile while optimizing for computational resources.
 
The ImageExporter module is responsible for using the Earth Engine Python API and builds upon the Python \href{https://geemap.org/}{GeeMap package} to implement the data extraction process. Rather than using the native JavaScript approach to interact with Google Earth Engine Code Editor, we devised a scalable and development-ready data extraction pipeline using Python. This module adeptly converts Python code into Javascript queries, dispatched to the Earth Engine API, facilitating the extraction of images that are subsequently stored in a designated Google Drive folder. By combining GEE’s robust data extraction capabilities with the fishnet’s systematic segmentation, this approach enhances scalability, standardization, and usability for large-scale, machine-learning-oriented spatial analyses.

To test our downstream urbanization prediction task, we needed to obtain a single image per year for each fishnet tile spanning from 2016 to 2022. This is generalizable to extracting images between any \textit{year\_start} and \textit{year\_end}. Although it's feasible to retrieve any image from a single time stamp within the year in question (say at the beginning or the end of the year), we chose to employ aggregation techniques to create a composite image. 

This process entails extracting multiple images for the same year, resulting in an image collection corresponding to multiple timestamps of one fishnet tile, and then applying pixel-level aggregation operations. Some of the available aggregation operations are average, minimum, maximum, median, etc. The advantage of retrieving an aggregated image over a single image lies in noise reduction, with noise stemming from various factors such as cloud cover or sensor noise. 

The assumption is that aggregation will lead to better signal due to its ability to smooth out short-term variations or anomalies. However, in highly dynamic environments, such as during rapid land-use changes or natural disasters, aggregation may inadvertently mask critical signals. To address this, our method can be adapted to account for such dynamics by employing several techniques, including:
\begin{itemize}
    \item Selective Aggregation: Aggregating over a subset of high-quality, temporally relevant images.
    \item Change Detection: Identifying temporal outliers or sudden deviations to emphasize areas experiencing rapid change.
    \item Dynamic Granularity: Adjusting the aggregation window based on the expected rate of change in the region.
\end{itemize}
If downstream machine learning models are used, additional strategies can be considered:
\begin{itemize}
    \item External Factors as Features: External factors can be incorporated as additional features or channels in the deep learning architecture, facilitating model learning.
    \item Avoiding Aggregation: A more granular time horizon can be used, feeding the model many more images and avoiding aggregation altogether.
\end{itemize}
Each of these approaches has trade-offs, and it is crucial to consider the specific use case when deciding which method to apply.

The Google Earth Engine offering has built-in functions for aggregating image collections into a single image, however, we opted to incorporate image aggregation in our ImageCorrector module to avoid needing to re-download images in case a different aggregation technique becomes necessary.

\begin{figure}[!t]
\centering
\subfloat[]{\includegraphics[width=1.8in]{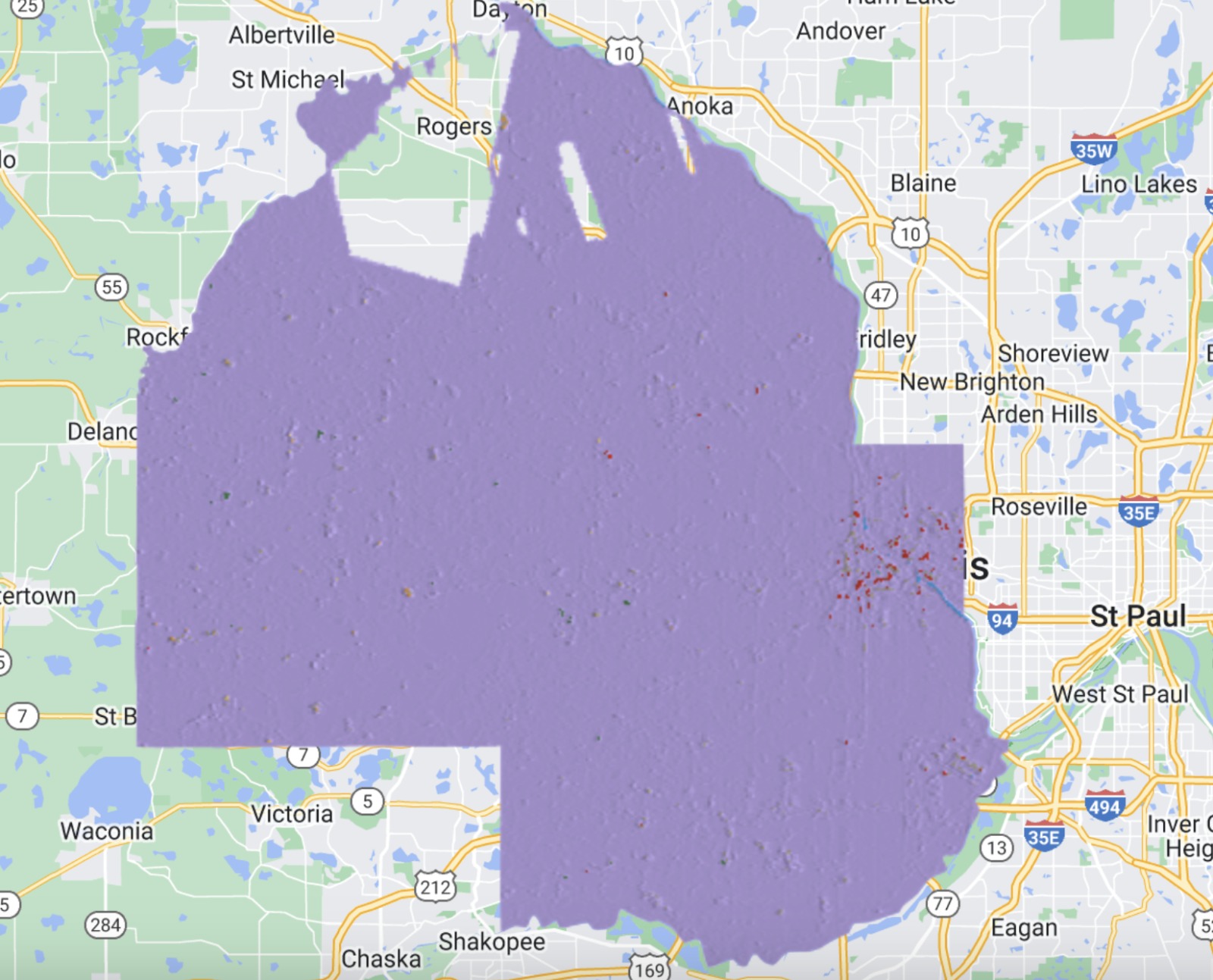}%
\label{fig_first_case}}
\hfil
\subfloat[]{\includegraphics[width=1.8in]{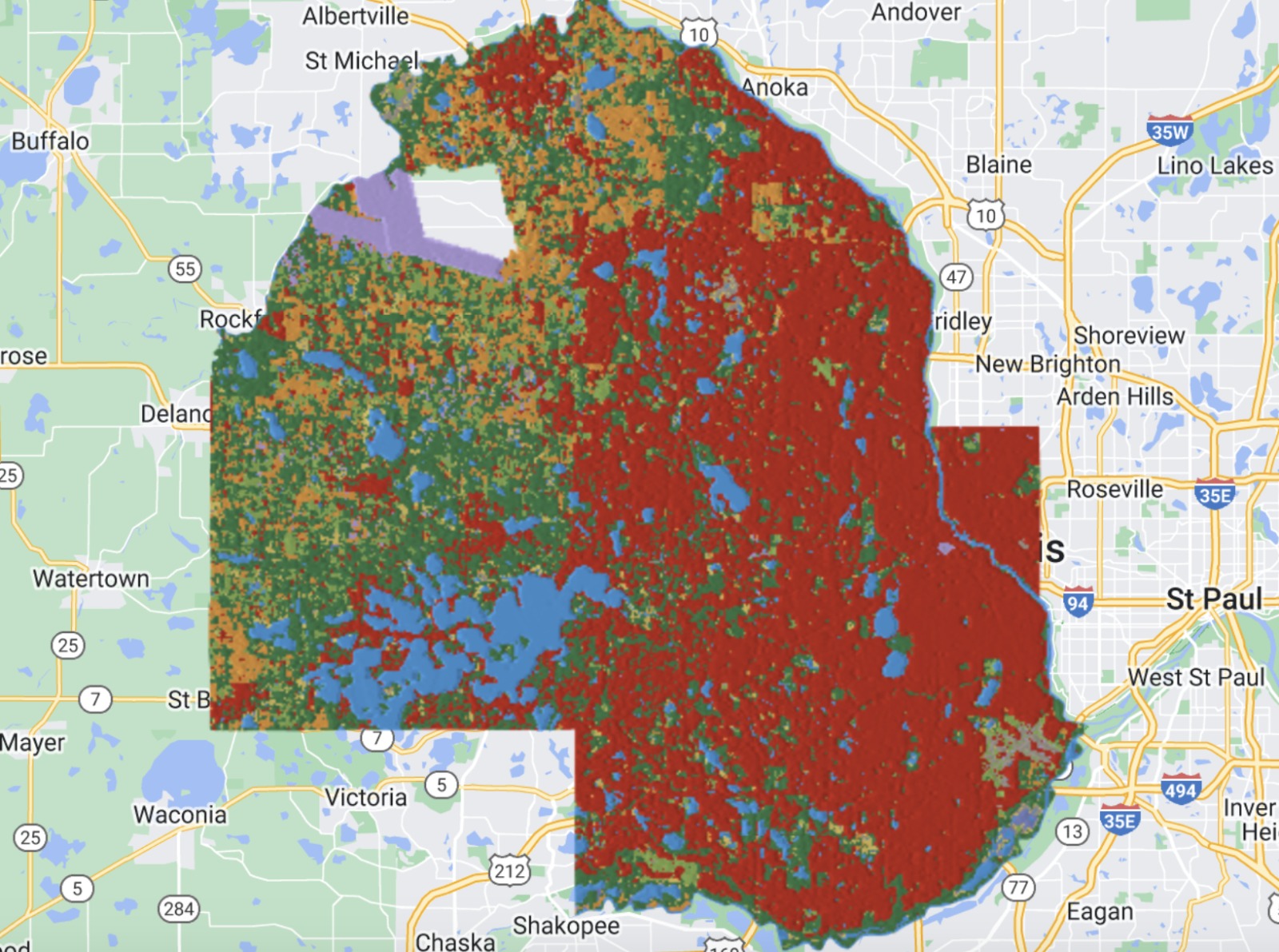}%
\label{fig_second_case}}
\hfil
\subfloat[]{\includegraphics[width=1.8in]{Pictures/hennepin2.jpeg}
\label{fig_third_case}}
\caption{The trio of images presented show the Dynamic World labels for Hennepin County, one of the United States' coldest areas, during various times in 2016. Figure \textbf{(a)} corresponds to the winter time period where the "Snow" label (purple) predominates, complicating land cover analysis. In contrast, Figure \textbf{(b)} corresponds to the summer time period, devoid of "Snow" labels but plagued with numerous gaps due to cloud coverage. Figure \textbf{(c)} depicts the annual average, featuring minimal snow and fewer missing values. The gaps in the summer data were filled using the annual averages.}
\label{fig:snow2}
\end{figure}

To address the challenges of seasonal variations, especially in cold regions where snow significantly affects land cover visibility in satellite imagery (Figure \ref{fig:snow2}), we only extracted and then aggregated images from summer months (June 1st to October 1st). This period is chosen because it offers a clearer view of labels in their most representative state, free from snow coverage that can mask the underlying land features.

\begin{figure}[!t]
\centering
\subfloat[]{\includegraphics[width=1.5in]{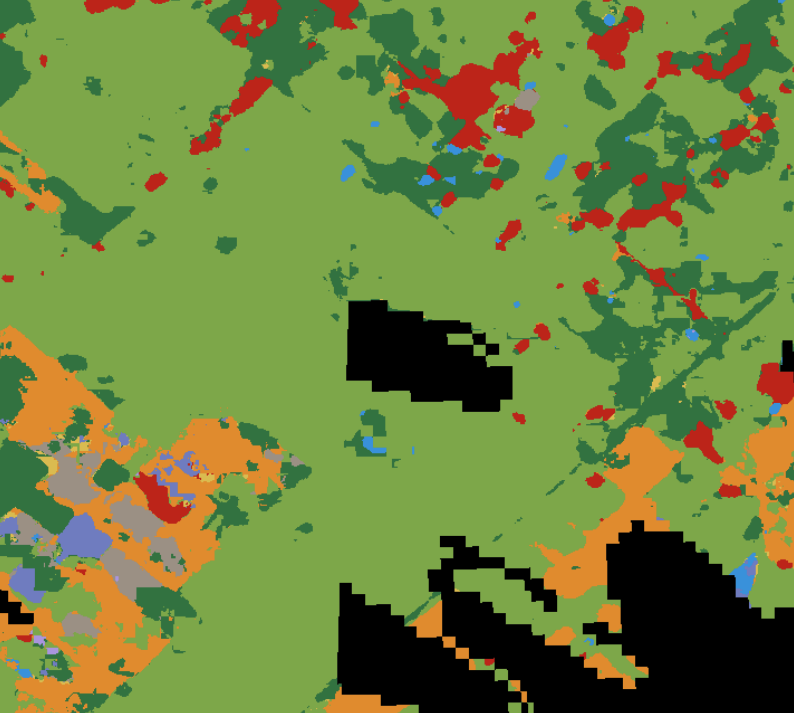}%
\label{fig_first_case_stains}}
\hfil
\subfloat[]{\includegraphics[width=1.5in]{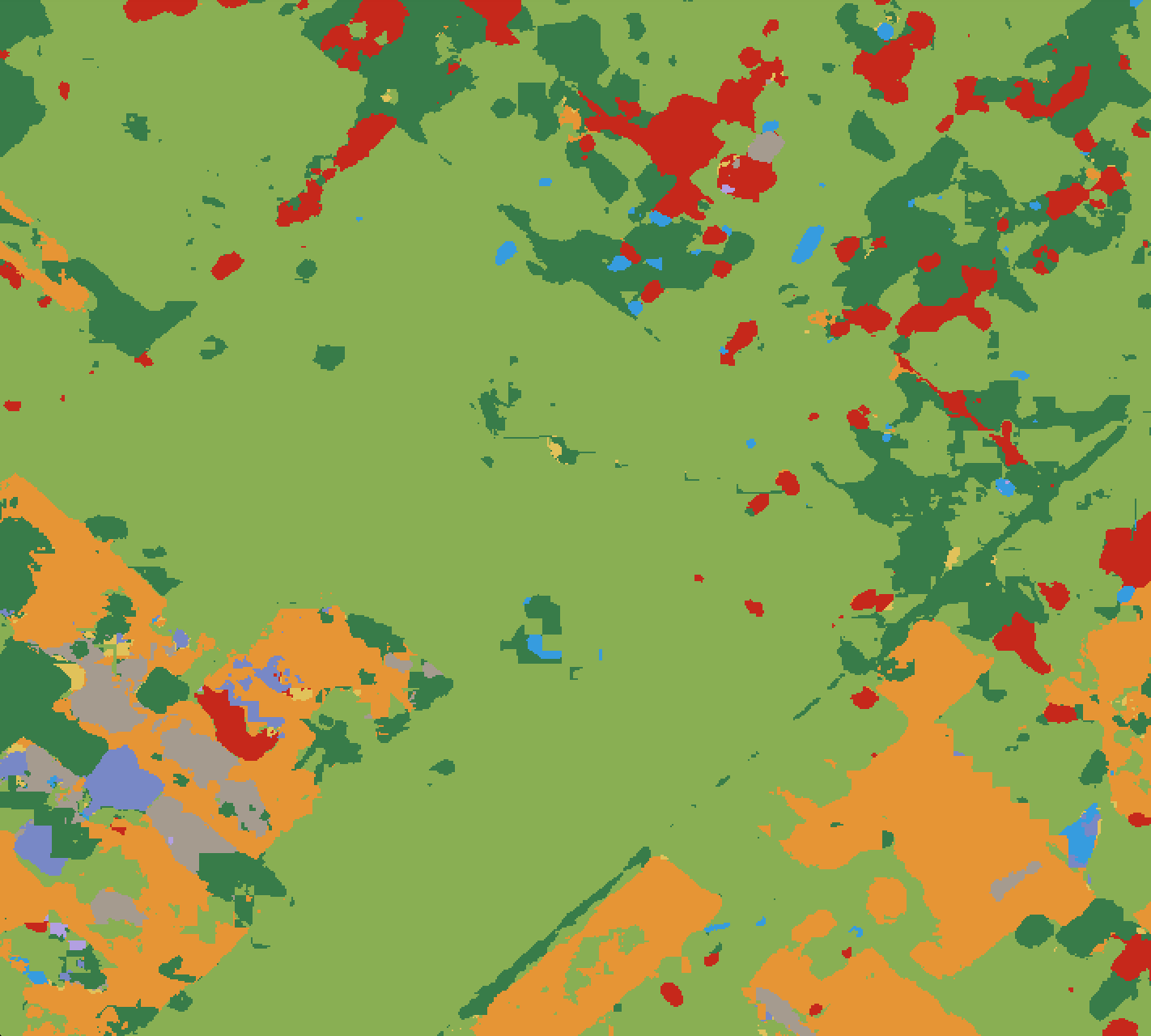}%
\label{fig_second_case_stains}}
\vfil
\subfloat[]{\includegraphics[width=1.5in]{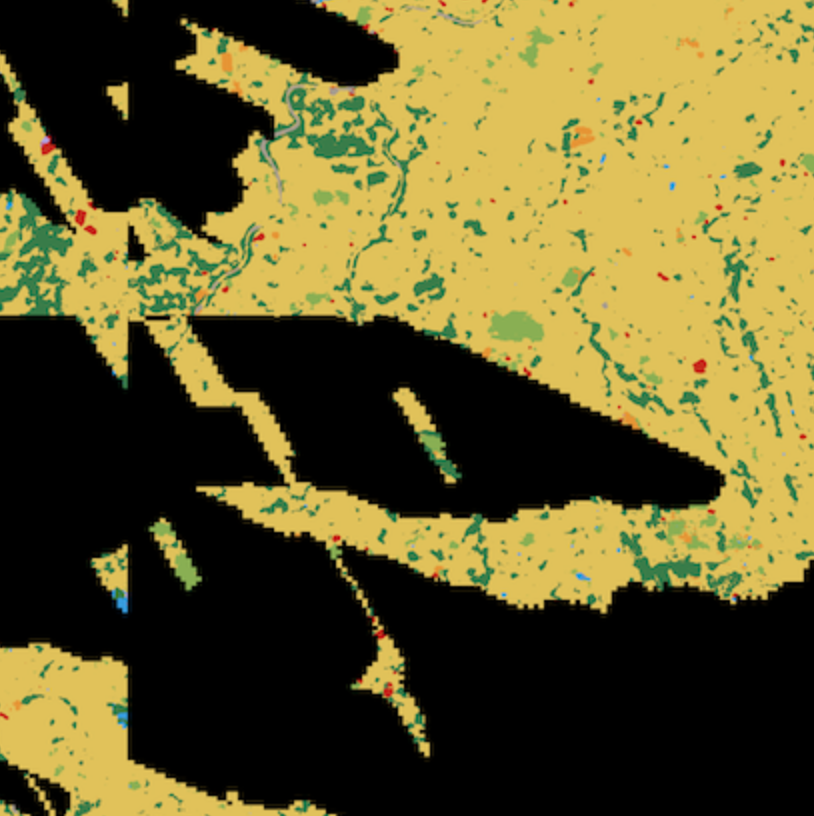}%
\label{fig_third_case_stains}}
\hfil
\subfloat[]{\includegraphics[width=1.5in]{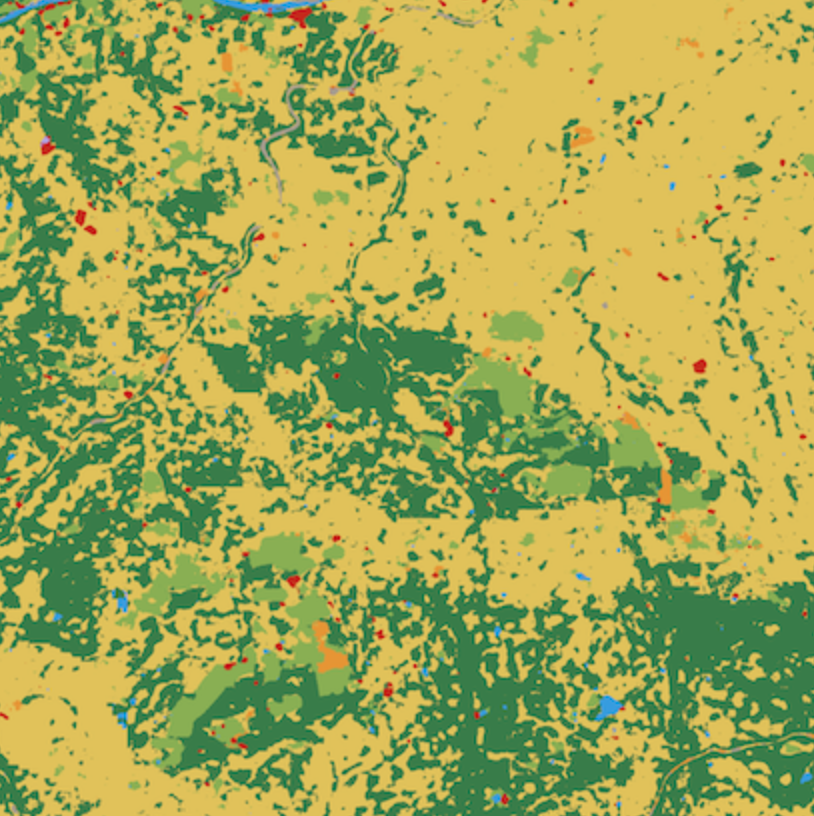}%
\label{fig_fourth_case_stains}}
\vfil
\subfloat[]{\includegraphics[width=1.5in]{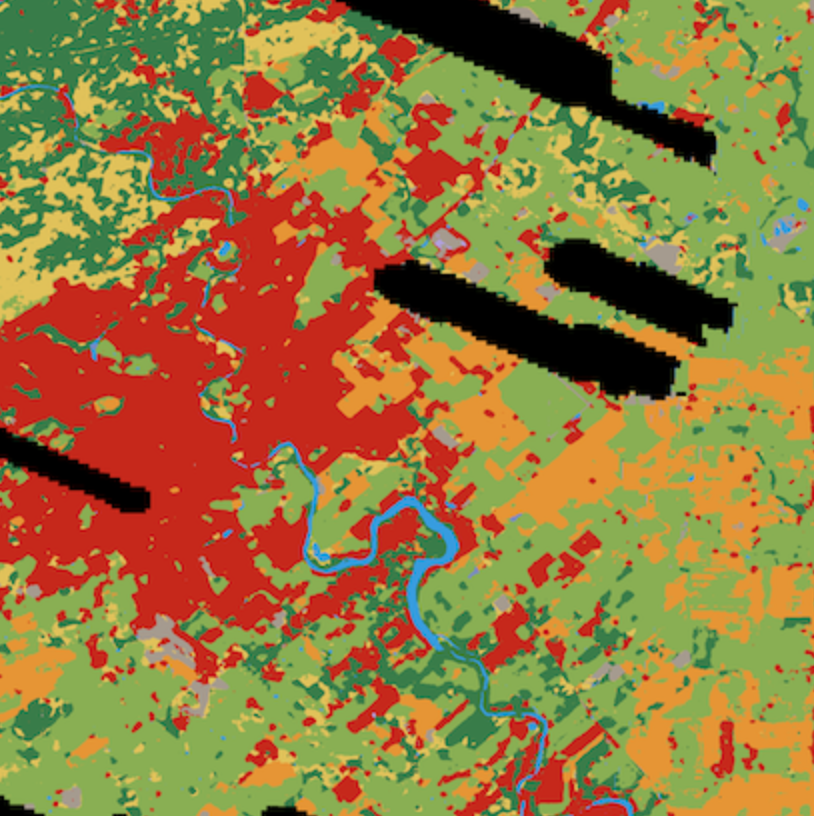}%
\label{fig_fifth_case_stains}}
\hfil
\subfloat[]{\includegraphics[width=1.5in]{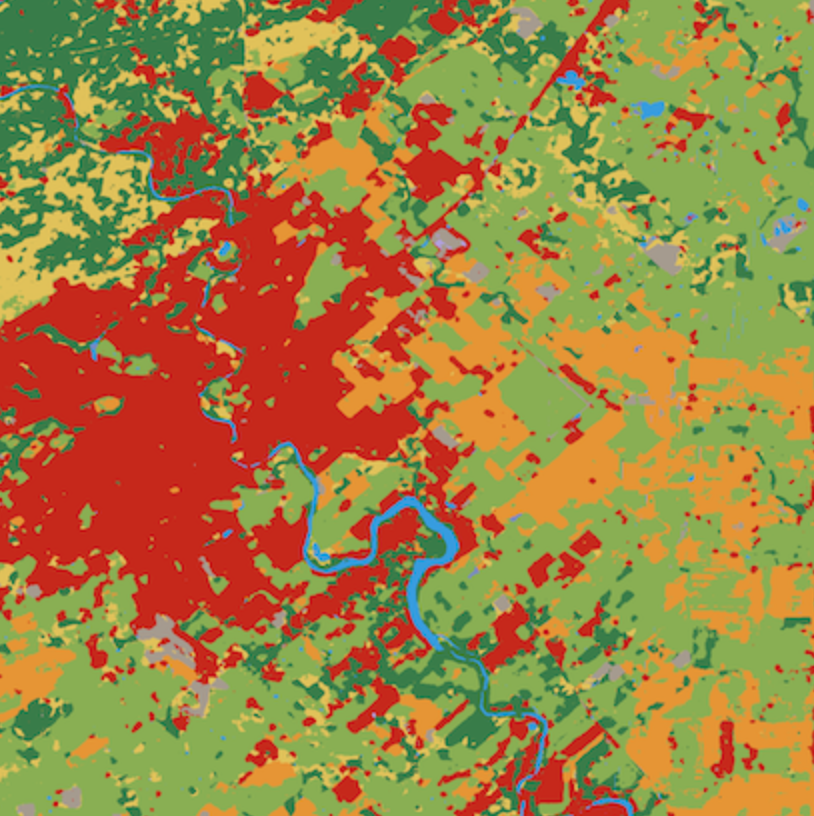}%
\label{fig_sixth_case_stains}}
\caption{Dynamic World exports: left before the correction; right after imputing missing pixels.}
\label{fig:correction2}
\end{figure}

While this strategy is successful in avoiding snow coverage and correctly capturing vegetation, it also encounters obstacles, such as cloud cover, which has particularly impacted data completeness in 2016 and 2017. To rectify instances of missing data caused by clouds, the ImageCorrector module includes imputation techniques, drawing missing pixels from images aggregated over a longer time period which are free from missingness. An example of this can be seen in Figure \ref{fig:correction2}.

% \begin{figure}[H]
%     \centering
%     \begin{subfigure}[b]{0.45\linewidth}
%         \centering
%         \includegraphics[width=\linewidth]{Pictures/black_stains.png}
%         \caption{Before correction}
%     \end{subfigure}
%     \hfill
%     \begin{subfigure}[b]{0.45\linewidth}
%         \centering
%         \includegraphics[width=\linewidth]{Pictures/no_stains.png}
%         \caption{After correction}
%     \end{subfigure}
%     \caption{Dynamic World exports: (a) before the correction and (b) after imputing missing pixels.}
%     \label{fig:correction}
% \end{figure}

Finally, the ImageProcessor class computes various aggregate metrics, such as the percentage of pixels with a certain value within each cell. These aggregate metrics are stored in tabular form, indexed by the unique fishnet tile identifier.

\section{Application Case Study and Results}

In this work we proposed a standardized extraction, pre-processing and representation pipeline for satellite images. Rather than performing this process in a one-off basis, our approach decreases the time and effort needed to obtain quality representations, freeing resources for data analysis or machine learning and obtaining quality results in a fraction of the time. To showcase our pipeline, we demonstrate how it can be used to build a competitive model for urbanization prediction.

Predicting the future state of a fishnet tile is complex because it involves understanding both temporal and spatial relationships in the input features as well as the dynamics of these relationships in areas with different growth profiles. The goal of this case study is to prove that with the data extraction and pre-processing pipeline previously described, even a complex task such as video prediction for urbanization forecasting is relatively simple to implement and achieves good results.

\begin{figure}[H]
\includegraphics[width=8.9 cm]{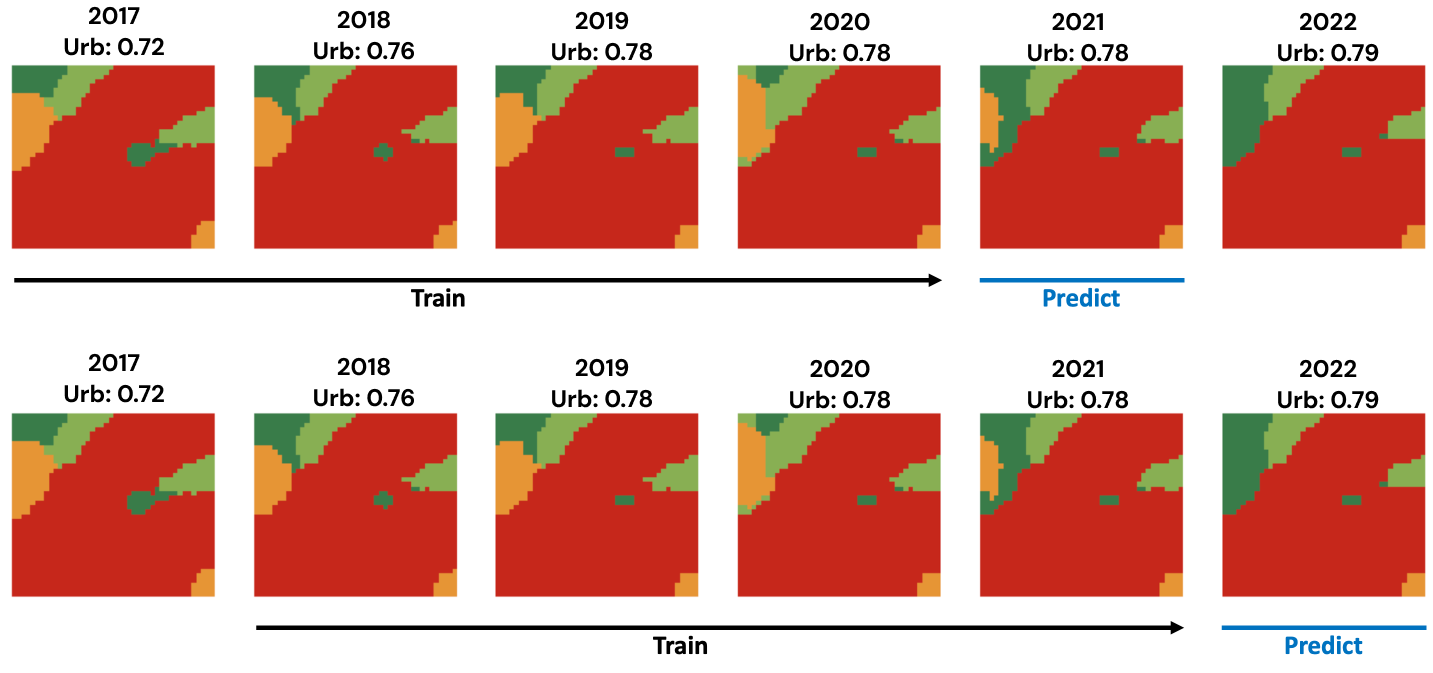}
\caption{Structure of ConvLSTM input and corresponding prediction. ConvLSTM architecture can be customized to predict either the whole next frame, or the aggregate value corresponding to that frame. In our case, we predict the aggregate value. }
\label{fig:train_predict_sequence}
\end{figure}

The model input is a sequence of satellite images from different years processed through our pipeline that can be conceptually regarded as a video, where each image serves as an individual frame (Figure \ref{fig:train_predict_sequence}). Through a sliding window mechanism, we train the model to predict the region's future state based on the previous N (here, N = 4) frames. 

\begin{figure*}[!t]
\centering
\includegraphics[width=0.8\textwidth]{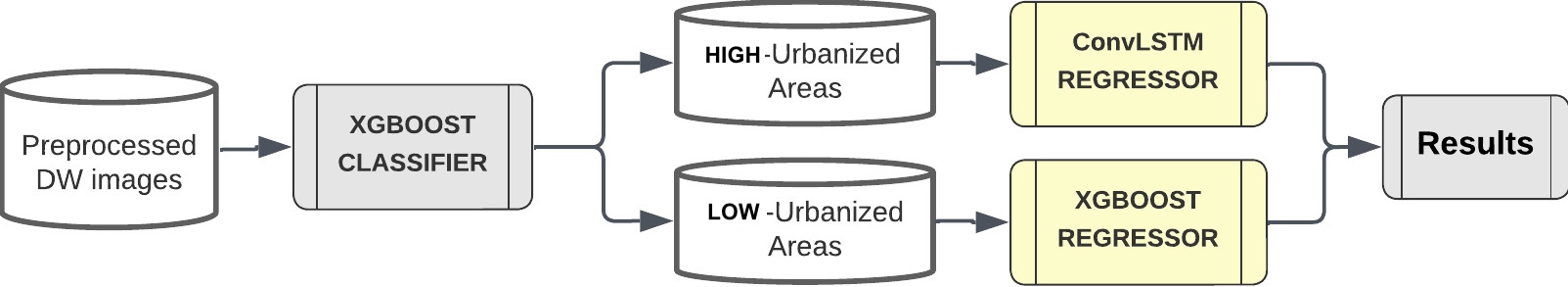} 
\caption{Workflow of an urbanization prediction model using XBCLM. Dynamic World images are classified by XGBoost to identify low and high urbanization areas. Regions with low urbanization are analyzed with an XGBoost Regressor and regions with high urbanization with a ConvLSTM Regressor, optimizing the prediction of urbanization patterns through tailored approaches for different development levels.}
\label{fig:architecture}
\end{figure*}

Figure \ref{fig:architecture} shows a diagram of our model architecture. First, the time series of images is fed into an XGBoost Classifier, trained to solve a binary classification problem, and the area/region is classified as actively urbanizing or passively stabilized. Actively urbanizing regions are fed into a ConvLSTM video prediction model that takes as input the sequence of images. Static regions are fed to an XGBoost Regressor model trained on tabular urbanization information extracted using the ImageProcessor method. More specifically, from each tile, the ImageProcessor method extracts the average number of urbanized pixels for different years to build a tabular dataset, as can be seen in \ref{fig:train_predict_sequence}. Both models predict the next state of the region in the sequence, represented by the proportion of urbanized pixels in the area.  Finally, the predictions from both branches of the pipeline are combined to produce a final urbanization prediction for the tile. We call XBCLM the overall model architecture.

The need for the initial XGBoost classifier stems from the lack of or minimal change over time in most fishnet tiles, a characteristic associated with all LULC satellite products. This poses a significant issue since deep learning models can get trapped in local minima, forecasting little to no change in the majority of fishnet tiles. The initial classifier prevents this issue, making XBCLM a powerful architecture. XGBoost was chosen for its known performance in terms of speed and accuracy; due to the case-study nature of this task, we did not assess different model choices.

As a comparison to XBCLM, we also trained the ConvLSTM with a customized loss function which puts a higher weight on urbanizing pixels to discourage the model from predicting no change. Instead of the typical MSE loss, as shown in equation \ref{equ:mse}, 
\begin{equation}
MSE = \frac{1}{N} \sum_{i=1}^{N} (y_i - \hat{y_i})^{2} 
\label{equ:mse}
\end{equation}

\noindent we implemented the custom Weighted MSE (WMSE), shown in equation \ref{equ:wmse}.

\begin{equation}
WMSE = \frac{1}{N} \sum_{i=1}^{N} (y_i - \hat{y_i})^{2} (1+W \cdot \alpha_{i})
\label{equ:wmse}
\end{equation}
where:
\begin{itemize}
\item ($W$) is a multiplier to model increased focus on urbanizing pixels (in our case W=100),
\item ($\alpha_{i} \in \{0,1\}$) represents whether element $i$ is an urbanizing pixel or not
\end{itemize}

\noindent While this slightly improved model performance, it did not give significantly better results, so XBCLM is the chosen approach.

Our study presents results derived from training on two distinct data modalities. XBCLM trained on satellite images is benchmarked against an XGBoost regressor which takes as input tabular features extracted from the tiles by the ImageProcessor method. 

Both models' output is a single value representing the tile's proportion of urbanized pixels in the future.
More formally, the XGBoost regressor takes as input a dataframe that lives in $R^{T \times Y}$, where $T$ denotes the total number of tiles and $Y$ represents the number of years used as features. The output is a vector in $R^{T}$. XBCLM takes as input $T$ sequences of $Y$ images, and also returns a vector in $R^{T}$.
We trained three ensemble and three XGBoost models, each predicting the urbanization proportion of a tile across different future time horizons: 1 year, 2 years, and 3 years in advance. More specifically, we use images from 2016 to 2019 to predict the state in 2020 (1 year in advance prediction), 2021 (2 years in advance) and 2022 (3 years in advance). 
This approach not only underscores the robustness of our proposed pre-processing pipeline in signal extraction, but also underscores its adaptability across different modalities.

To prevent data leakage between training, validation, and testing phases, we selected entirely distinct regions for each set. Specifically, we utilized the state of Texas for training, Georgia for validation, and Alabama for testing. While we acknowledge that testing the models on additional regions would contribute to more robust results, the primary focus of this paper is on the data pre-processing pipeline rather than the machine learning model itself. Therefore, we limited our testing to these regions.

\begin{figure}[H]
\includegraphics[width=8.9 cm]{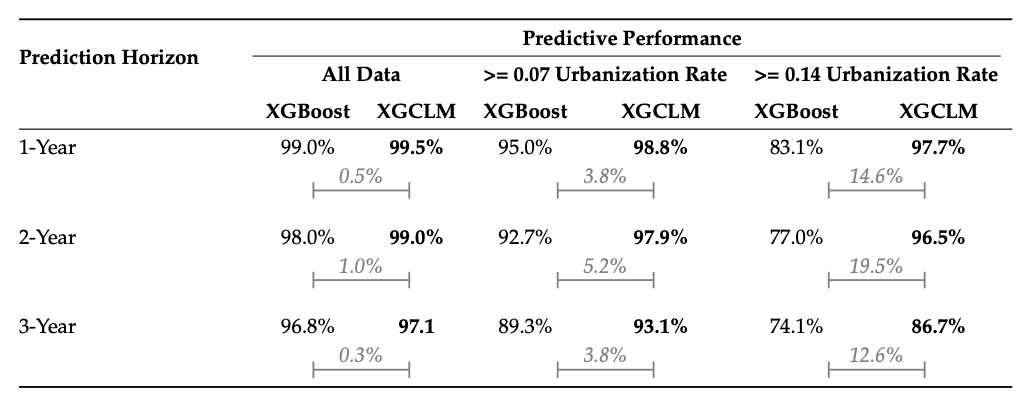} 
\caption{Model results comparison, bar represents the increase in performance.}
\label{fig:results}
\end{figure}

The results of the XGBoost regressor model are compared to those of the XGCLM model framework in Table \ref{fig:results}.

The suggested pre-processing pipeline enables the extraction of data from various modalities, facilitating the training of both XGCLM and XGBoost Regressor models. As seen in the table, the XGCLM model outperforms the XGBoost Regressor at all time horizons. Interestingly, the higher the urbanization rate, the better the performance of the XGCLM model compared to the XGBoost Regresor. This demonstrates that video prediction is able to capture the dynamics of the data at different rates of urbanization.

It is important to note that $R^2$ tends to be very high even for the XGBoost model, especially in areas that do not urbanize quickly. The reason is that, at the tile level, change happens rarely and slowly. 

To address the evaluation of our pipeline with traditional approaches, we implemented a Cellular Automata (CA) model, which is widely used in the literature for spatio-temporal forecasting in remote sensing. Cellular Automata models consist of a grid of cells that evolve through discrete time steps according to simple, local rules based on the states of neighboring cells, leading to complex global patterns. This CA model was trained using the output of our preprocessing pipeline. Due to its pixel-by-pixel nature, the CA model is computationally intensive and was trained on a limited dataset, resulting in modest performance metrics (R²: 0.42). However, it is important to note that computations for CA can be parallelized, which would result in significant speed-ups. In our case, we did not implement such optimizations as the primary focus of this paper was to demonstrate the flexibility of our preprocessing pipeline rather than optimizing the CA approach. While these results do not match the performance of the XGBoost and XGCLM models, they demonstrate the pipeline's adaptability to traditional approaches like CA.

This highlights the pipeline's flexibility, enabling the extraction of data from various modalities and supporting both traditional and advanced modeling techniques. Depending on the training complexity and performance needed, one can opt to train on either tabular data or images directly. Our work accommodates many different modeling approaches, and the impressive performance of both XGBoost and XGCLM models highlights the quality and stability of our pre-processing pipeline.

\section{Interesting Research Avenues}

Our current pipeline is primarily structured to handle Dynamic World satellite images. In future studies, we hope to include other LULC datasets. Another potential enhancement to this study would be to expand the pipeline's capabilities to seamlessly process various other types of satellite imagery, each encoding different information. In this case, the output would contain multiple customizable information layers, providing a more comprehensive and tailored analysis. 

For instance, in our urbanization prediction case study, incorporating nighttime lights imagery could prove valuable in identifying urban areas and monitoring their growth patterns. Other relevant data sources could include surface temperature measurements, air quality indices, cloud coverage maps, or wildlife distribution data. By integrating these diverse information layers, our analysis could gain deeper insights and produce more accurate and holistic predictions.

Additionally, we acknowledge the role of external factors such as new regulations, city plans, or unforeseen socio-economic developments in shaping land cover dynamics. While our work focuses on historical data as a proof-of-concept case study, the proposed framework is inherently flexible and extensible, allowing for the inclusion of exogenous features that can represent these external factors. 
For example, the neural network architecture can accommodate additional information as input features or channels. Data such as proximity to major roads, distance to city centers, land use restrictions (e.g., private property, protected areas, or reservoirs), or planned infrastructure developments can be encoded into the model. These features could be incorporated as auxiliary inputs or directly encoded into the spatial grid as additional channels. For instance, new urban development regulations might be represented as categorical features at the tile level, indicating development eligibility or restrictions. Similarly, planned city infrastructure, such as roads or parks, could be encoded with specific values to guide the model’s predictions.
Since it is complex to manually obtain and encode such information for large regions, large language models (LLMs) could be leveraged to analyze regulatory documents, city planning reports, and socio-economic updates to determine which geographic regions may be impacted. These insights could then be encoded as features for the forecasting model, enabling it to dynamically adapt to evolving conditions.

Another promising research avenue would be to test this approach across a diverse range of Dynamic World labels and study topics such as deforestation, annual snow patterns in the Alps, and agricultural land use dynamics. By applying our methodology to these varied contexts, we could broaden its applicability and evaluate its efficacy in addressing diverse environmental and land-use challenges. Such case studies would not only validate the model's versatility but also provide valuable insights into different geographical and ecological phenomena, contributing to a more comprehensive understanding of the Earth's dynamic systems.

\section{Conclusion} % \section{Discussion}

Our study introduces a standardized pipeline tailored for the extraction, pre-processing, and representation of LULC data sourced from the Dynamic World dataset. As proof of the efficacy of our pre-processing pipeline, we present a predictive model trained on images pre-processed through our pipeline. Results on urbanization patterns show that this model is capable of forecasting future states of LULC labels. 

One of the principal strengths of our pipeline lies in its accessibility and user-friendliness. By automating the extraction and pre-processing of satellite images, our approach requires minimal user input, streamlining the generation of analysis-ready datasets. It also ensures standardization and simplifies benchmarking for downstream models, addressing the limitations of heterogeneous custom pipelines. Moreover, our pipeline facilitates the representation of satellite imagery in tabular formats, enhancing versatility for subsequent analytical tasks. The seamless integration between image and tabular data enables researchers to approach analysis and prediction tasks through diverse modalities, catering to the specific requirements of each task.

Addressing the inherent challenge of predicting LULC changes, we devised a two-stage modeling approach. First, our model distinguishes between regions with high urbanization and those with low urbanization using a classification model. Depending on its classification prediction, the information is fed into one of the two potential predictive models, each responsible for estimating the magnitude of urbanization change. Our model is able to predict future urbanization with high accuracy, outperforming the baseline significantly in areas where the rate of urbanization is high.

By reducing the overhead associated with data pre-processing, our model unlocks the rich information contained in the Dynamic World dataset and facilitates the exploration of environmental changes, population movements, and weather phenomena. We advocate for the broader adoption of our extraction pipeline to foster advancements in remote sensing research and its application across various domains.

\section*{Acknowledgments}
We extend our gratitude to the MIT Sloan School of Management for their support through the Master of Business Analytics (MBAn) capstone project, which provided essential resources for this study. Additionally, we would like to acknowledge JP Morgan for their contribution in supplying resources that facilitated our research.

\section*{Data Availability}
For this study, we exclusively utilized the \href{https://dynamicworld.app/}{Dynamic World dataset}. No new data were created during our research. The Dynamic World dataset, which supports our findings, is publicly available and can be accessed through Google's Earth Engine platform. This dataset provides global land cover classification at high temporal resolution, allowing for detailed analysis of land use and land cover changes over time. Further information on accessing and using the Dynamic World dataset can be found at the official Google Earth Engine data catalog.

\section*{Disclaimer}
This paper was prepared for informational purposes in part by the Artificial Intelligence Research group of JPMorgan Chase \& Co. and its affiliates ("JP Morgan'') and is not a product of the Research Department of JP Morgan. JP Morgan makes no representation and warranty whatsoever and disclaims all liability, for the completeness, accuracy or reliability of the information contained herein. This document is not intended as investment research or investment advice, or a recommendation, offer or solicitation for the purchase or sale of any security, financial instrument, financial product or service, or to be used in any way for evaluating the merits of participating in any transaction, and shall not constitute a solicitation under any jurisdiction or to any person, if such solicitation under such jurisdiction or to such person would be unlawful.

\section*{Abbreviations}{
The following abbreviations are used in this manuscript:

\begin{table}[H]
\centering
\begin{tabular}{|c||c|}
\hline
DW & Dynamic World \\
\hline
convLSTM & Convolutional Long Short-Term Memory \\
\hline
XGBoost & eXtreme Gradient Boosting \\
\hline
LULC & Land Use and Land Cover \\
\hline
UEP & Urban Expansion Prediction \\
\hline
XGCLM & XGBoost-ConvLSTM Model \\
\hline
\end{tabular}
\end{table}

\bibliographystyle{IEEEtran}
\bibliography{refs}

\vfill

\end{document}